%% file: acl2023.tex
\pdfoutput=1

\documentclass[11pt]{article}

\usepackage{ACL2023}

\usepackage{times}
\usepackage{latexsym}
\usepackage{booktabs}
\usepackage[T1]{fontenc}
\usepackage[utf8]{inputenc}
\usepackage{microtype}
\usepackage{inconsolata}
\usepackage{soul}
\usepackage{cleveref}
\usepackage{xspace}
\usepackage{enumitem}
\usepackage{booktabs}
\usepackage{multirow}
\usepackage{float}
\usepackage{xcolor}
\usepackage[disable]{todonotes}

\newcommand{\ie}{i.e.,~}
\newcommand{\eg}{e.g.,~}

\newcommand{\example}[1]{{``\emph{#1}''}\normalsize}

\title{Generate-then-Retrieve: Intent-Aware FAQ Retrieval in Product Search}

\author{Zhiyu Chen~~~~~~Jason Choi~~~~~~Besnik Fetahu~~~~~~Oleg Rokhlenko~~~~~~Shervin Malmasi \\
  Amazon.com, Inc.~~~~ Seattle, WA, USA \\
{\{\texttt{zhiyuche,chojson,besnikf,olegro,malmasi\}@amazon.com}}}

\begin{document}
\maketitle
\begin{abstract}
Customers interacting with product search engines are increasingly formulating information-seeking queries.
Frequently Asked Question (FAQ) retrieval aims to retrieve common question-answer pairs for a user query with question intent. Integrating FAQ retrieval in product search can not only empower users to make more informed purchase decisions, but also enhance user retention through efficient post-purchase support. Determining when an FAQ entry can satisfy a user's information need within product search, without disrupting their shopping experience, represents an important challenge.
We propose an intent-aware FAQ retrieval system consisting of (1) an intent classifier that predicts when a user's information need can be answered by an FAQ; (2) a reformulation model that rewrites a query into a natural question. 
Offline evaluation demonstrates that our approach improves Hit@1 by 13\% on retrieving ground-truth FAQs, while reducing latency by 95\% compared to baseline systems. These improvements are further validated by real user feedback, where 71\% of displayed FAQs on top of product search results received explicit positive user feedback. Overall, our findings show promising directions for integrating FAQ retrieval into product search at scale.
\end{abstract}

\input{intro2}

\input{related_work}

\input{method}
\input{exp}
\input{conclusion}

\input{limit}
\bibliography{custom}
\bibliographystyle{acl_natbib}
\end{document}

%% file: intro2.tex
\section{Introduction}
\label{sec:intro}

Product search engines help users find relevant products across large product catalogues and generate sales revenue for e-commerce companies \citep{grover2001commerce}. 
While such engines are primarily designed to handle keyword searches for products, customer behavior has been changing with an
increase in users asking information-seeking service or product related questions~\cite{carmel2018product,gao2019product}.
However, most product search engines are not effective at handling non-product search related queries (\eg ``return a package"). Providing correct answers to these Frequently Asked Questions (FAQs)~\citep{gupta2019faq, mass2020unsupervised} is essential to provide a positive pre- and post-purchase experience, which can lead to improved user retention and trust.

Product search and FAQ retrieval are typically powered by independent retrieval systems. This separation is often due to the challenges in combining multiple answering sources (e.g. product details and FAQs) into a holistic retrieval application~\cite{park-etal-2015-question,christmann2022conversational}. 
Furthermore, determining what answering source can satisfy the user's information need is challenging to perform at scale.

Hence, e-commerce websites tend to isolate FAQ search functionality from product search. For example, Apple offers vertical search where users are required to navigate among different tabs (e.g., product search, support, store location).\footnote{\tiny\url{https://www.apple.com/us/search/apple-tv-bluetooth}}
Such designs require users to navigate multiple links, which can lead to increased user effort and unsatisfactory shopping experiences~\citep{siraj2020characteristics,nain2021tourism,su2018user}. 
Therefore, we propose to integrate FAQ retrieval into a product search engine, so that users can search products and access FAQs seamlessly from an unified search interface.

\begin{figure}[t]
    \centering
    \includegraphics[height=120pt]{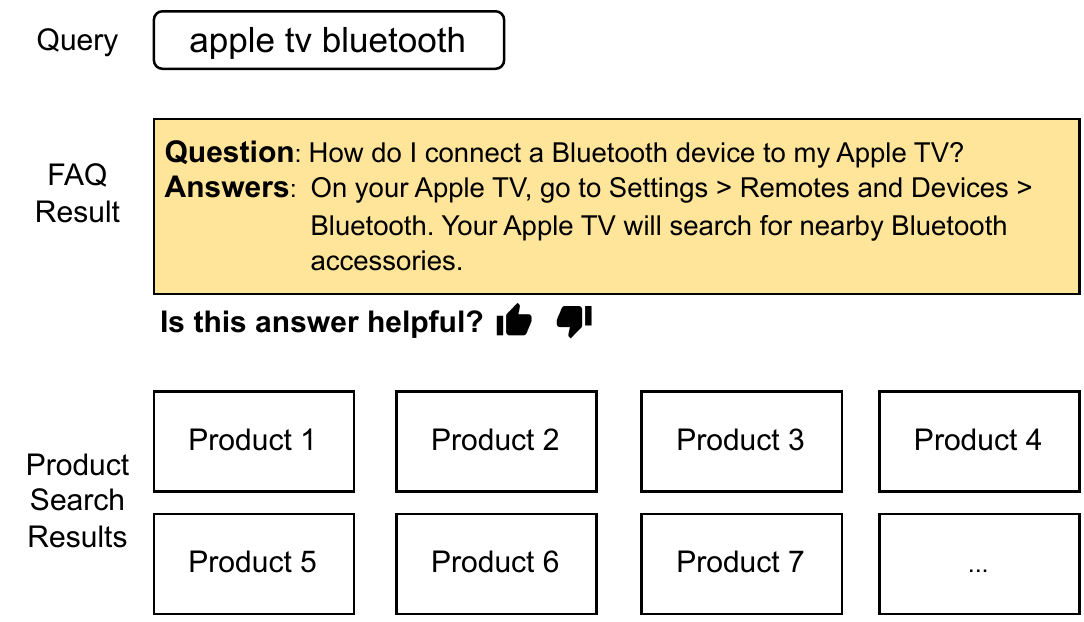}
    \caption{Our proposed aggregated search interface that jointly displays top-1 FAQ and product search results for queries with question intent.}
    \label{fig:faq_example}
\end{figure}

A potential solution is \textit{aggregated search}, which refers to the task of searching and assembling information from a variety of sources and presenting them in a single unified interface \citep{murdock2008workshop,wan2008aggravated,lu2012simulating}.
The main challenge here lies in determining \textbf{when} and \textbf{how} information from multiple \textit{verticals} should be presented effectively and efficiently.

\paragraph{When to Show FAQ Results?}
\label{sec:c1}
Query intent is inherently ambiguous~\cite{krovetz1992lexical,song2007identifying,sanderson2008ambiguous}. Figure~\ref{fig:faq_example} illustrates an example where the users can use the same query \example{apple tv bluetooth} to retrieve products, or find information about Bluetooth connectivity. In the latter case, the query is intended to express the question, \example{How do I connect a Bluetooth device to my Apple TV?}. 
We define a query that can be answered by an FAQ entry as having \textit{question intent}. It is important to note that a query with question intent may also have product search intent, as queries can be inherently ambiguous.

Determining when a user query can be answered by either \emph{product} or \emph{FAQ search} is tightly coupled with predicting the user's information need. Due to query ambiguity, displaying answers from FAQ sources for all searches causes high false positive rates, due to the lack of question intent by the user.

Our analysis (cf. \S\ref{sec:intent_rs}) from a leading e-commerce site shows that if FAQs would be shown to all queries, 98\% of the FAQs would be irrelevant to user's needs. As users mostly use product search for shopping, injecting FAQ results that are irrelevant or not needed causes significant friction in the user experience.
Furthermore, performing FAQ retrieval for every query is inefficient \cite{tsur2016identifying} since only a small portion of traffic has question intent \cite{white2015questions}.\footnote{In the case of the Bing search engine only 10\% of queries were shown to have question intent.} While we cannot disclose the intent distribution of our data for reasons of confidentiality, question intent represent a minor portion of the overall query traffic.

To address the problems above, we train an intent classifier that distinguishes when a query has question intent, and thus, can be answered by an FAQ source. In mixed retrieval scenarios, this allows us to trigger FAQ retrieval and show FAQ results only for queries with question intent, causing less friction for users.
In terms of efficiency, running FAQ retrieval only on question intent queries significantly improves latency. 
Our experiments validate that deploying intent classifier brings a 95\% latency reduction compared to baselines without the intent classifier. Lastly, we demonstrate that existing techniques such as upsampling are enough to achieve satisfactory performance in classifying question intent in imbalanced traffic.

\paragraph{How to Show FAQ Results?} When a search query has question intent, an interface that jointly displays product search and FAQ retrieval results is required. As in prior work on aggregating web search results \citep{diaz2009integration}, we integrate the top-1 FAQ result alongside product search results, as illustrated in Figure~\ref{fig:faq_example}, for the following two reasons.

First, since product search is the core functionality of e-commerce search engines, majority of the space is dedicated to the ranked product list. If we consider additional modalities such as mobile search, space constraints are even greater. Displaying more FAQ results comes at the cost of reducing the number of product results, which can lead to reduced revenue \cite{feng2007implementing}.
Second, compared to product search, where users are required to compare multiple options, question intents generally require less exploration since users already have a specific request in their mind.

Given the above reasons, we need to optimize the FAQ retrieval system for \textbf{high precision} at the top ranks (\ie Hit$@1$). Queries are usually short and consists of several keywords. To achieve a high precision for FAQ retrieval, we propose to rewrite a query with question intent into a more specific natural language question. This rewriting process aims to make the queries semantically and syntactically more similar to the questions found in FAQs than the original queries, inspired by previous studies of query reformulation~\cite{zhao_automatically_2011, zheng_k2q_2011, yu2020few}. %
Our experiments validate that through query reformulation we can achieve significantly higher accuracy in retrieving ground-truth question at first rank (Hit$@1$) with more than 13\% improvement when compared to using original queries for FAQ retrieval. %

\paragraph{Contributions}
We summarize our contributions in this paper as follows:
\begin{itemize}[leftmargin=*]
\itemsep0em
    \item To our best knowledge, this is the first work to integrate FAQ retrieval and product search at scale.
    \item Our proposed intent-aware FAQ retrieval approach is a practical solution that significantly improves performance compared to baseline methods. Our approach achieves a 13\% higher precision in the top-ranked results (Hit$@1$) and is 95\% more efficient than baseline methods.
    \item To evaluate our design from a user's perspective, we reviewed feedback from users who interacted with a deployed version of our system. Results showed that 71\% of the rendered top-1 FAQ results (at the query level) received explicit positive customer feedback when displayed along with product search results.
\end{itemize}

%% file: related_work.tex
\begin{figure*}[h]
    \centering
    \includegraphics[width=2\columnwidth]{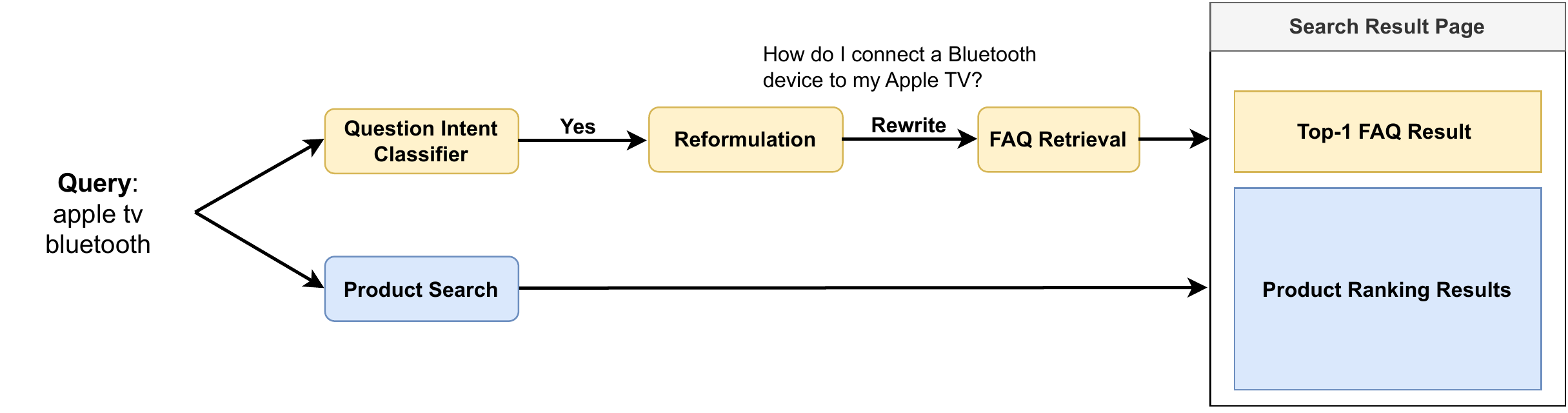}%
    \caption{An overview of our proposed intent-aware FAQ retrieval approach. While product search is performed by default, FAQ retrieval is triggered only for queries with question intent.}
    \label{fig:pipeline}
\end{figure*}

\section{Related Work}\label{sec:related_work}  

\noindent \textbf{FAQ retrieval.} The problem of FAQ retrieval has been extensively studied. Early methods~\cite{whitehead1995auto,sneiders1999automated} rely on exact keyword matching in FAQs. \citet{karan2013frequently} propose to derive lexical features such as n-gram overlap and TF-IDF similarity from a query-FAQ pair, and use these features to train a SVM model to classify whether the query is relevant to an FAQ. \citet{karan2016faqir} combine the scores from BM25, and a classical vector space model to rank the FAQ based on its semantic similarity to the query. More recently, deep learning methods have been applied to FAQ retrieval. \citet{gupta2019faq} adapt a sentence matching model~\cite{ijcai2017p579} based on bidirectional LSTM to aggregate question-to-query and question-to-answer similarities. Building on the success of BERT in NLP tasks, \citet{sakata2019faq} and \citet{mass2020unsupervised} have adopted BERT to rank answers or questions of FAQs using user queries.

However, all of the experiments from early work assume that the FAQ retrieval system is deployed independently and all input queries have a question intent. In our work, we evaluate our proposed search interface in a more realistic setting by simulating search traffic queries with both product search and questions intents (with significant imbalance). We argue that this setting is more suitable for studying the benefits of aggregated search interfaces \citep{murdock2008workshop} in large-scale e-commerce businesses.

\noindent \textbf{Keyword-to-Question Generation.}
Generating questions from search queries is an example of query rewriting that is first used in community-based question answering websites (\eg Yahoo! Answers, and Quora) to retrieve related questions. \citet{zhao_automatically_2011} propose a template-based method to rewrite keyword-based queries into questions by first extracting a set of query-question pairs from search engine logs. Then for each input query, the most relevant templates are retrieved to generate questions. A similar method is proposed by \citet{zheng_k2q_2011}, but the difference is allowing users to refine the generated question with generated refinement keywords. \citet{ding_generating_2018} use a statistical model to synthesize keyword-question pairs which are then used to train a neural model~\cite{gu-etal-2016-incorporating}. However, the synthesized queries are noisy and additional filtering mechanism has to be used to improve performance. 
Recently, \citet{iovine-etal-2022-cyclekqr} proposed a bidirectional keyword-question rewriting model that leverages non-parallel data through cycle training. Their experiments showed that sequence-to-sequence text generation models can perform the keyword-to-question task with high accuracy, and improve retrieval results in various scenarios.

Inspired by earlier keyword-to-question generation approaches, we utilize the state-of-the-art generation models to reformulate keyword queries into questions for FAQ retrieval. Experiments show that our reformulation model trained from human annotated query-question pairs significantly improves Hit@1 by 13\% compared to using original query.

%% file: method.tex
\section{Method}
\label{sec:method}

Our intent-aware FAQ retrieval approach, shown in Figure \ref{fig:pipeline}, consists of two main components: (1) an intent classifier that takes a user query as input and determines whether it has question intent and can be answered by an FAQ; (2) a query reformulation model which reformulates queries with question intent into a natural language question that is used for FAQ retrieval. Regardless of query intent, the product search is always performed.\footnote{Product search is beyond the scope of our work.}

\subsection{Question Intent Classification}
\label{sec:method_intent_classifier}

Unlike prior work on FAQ retrieval, we do not assume that all input queries have question intent. Instead, we train a binary intent classifier that takes an input query and predict its intent into: (1) non-question intent; (2) question intent. The intent classifier corresponds to a fine-tuned RoBERTa model~\cite{liu2019roberta} trained for the binary classification task. To handle class imbalance, we oversample the minority class (question intent) to approximate a balanced class distribution.

\subsection{Query Reformulation}
Once a query is classified as having question intent, the query is reformulated into a natural language question. We train a sequence-to-sequence Transformer model~\cite{vaswani2017attention} that reformulates the query into a question. %
The natural language question is used for FAQ retrieval, which are discussed in details in \S\ref{sec:data_reformulation}. The assumption behind our method is that generated questions are syntactically closer to FAQ questions than the original keyword queries, which can bring additional improvements in FAQ retrieval.

\subsection{Proposed Intent-Aware FAQ Retrieval}

As illustrated in Figure~\ref{fig:pipeline}, once a query is identified with question intent and is reformulated into a natural language question, the FAQ retrieval component takes it as input and returns %
the top-1 FAQ result. If a query does not contain a question intent, we do not initiate the FAQ retrieval process and only return product search results.

Our retrieval component ranks FAQ results solely based on questions, without utilizing the associated answers. The rationale behind this is that a well-reformulated question should closely match the ground-truth question of an FAQ entry, allowing a simple ranking component to accurately position the correct FAQ at the top position using only the question as input.
We claim that our intent-aware FAQ approach can satisfy users' information needs whether they are looking for FAQs, qualified products, or both, and therefore provides a more convenient pre-purchase and post-purchase experience.

%% file: exp.tex
\section{Experimental Setup}
In this section, we first discuss datasets used, followed by implementation details of retrieval baselines, experimental settings and evaluation metrics.%

\subsection{Datasets}
\label{sec:data}

\paragraph{Intent Classification Dataset} 
\label{sec:data_intent_classifier}
As the majority of search queries issued to e-commerce websites are focused on product search, annotations on random samples of queries yield only a tiny fraction of queries with question intent. This is not suitable for our needs, as we require a training set that includes a balanced distribution of intents. This challenge is addressed by applying several cycles of (1) training an intent classifier; (2) generating predictions on a set of unseen queries; (3) select question-intent queries with high probability (>0.9) for the next annotation; (4) correct model predictions through human annotations.

For the first cycle, when human annotations were not yet available, we utilized an existing keyword extraction algorithm (RAKE) \cite{rose2010automatic} on a publicly available product question corpus~\cite{rozen2021answering} to generate pairs of (question, query with question intent). For example, given a product question "How do I connect a Bluetooth device to my Apple TV" , we can extract keywords "connect Bluetooth device Apple TV" as the corresponding query with question intent. For queries with shopping intent, we randomly sampled queries from our traffic and applied a simple filter (e.g., removing queries that start with a question word). Based on these initial training samples, we prototyped our first model and improved it through subsequent human annotations. More details on intent classifier training are summarized in \S \ref{subsec:ours}.

Overall, our annotations resulted in 18,972 queries including 5,562 question-intent queries and 13,410 non-question queries. We allocate 50\% of the data as the training set, 25\% as the validation set, and the remaining 25\% as the testing set. Note that since there are multiple cycles of training and annotation, the distribution of intent in this set does not reflect the distribution of the real traffic.

\paragraph{Query Reformulation Dataset}
\label{sec:data_reformulation}
From the intent classification dataset, we randomly sample 1,500 question intent queries from the training set and 500 question intent queries from the testing set. For each query, we ask annotators to reformulate it into a natural language question. The resulting 1,500 query-question pairs form the training set is used for training reformulation models, while the remaining 500 pairs are divided evenly into validation and testing sets.

\paragraph{FAQ Corpus}
\label{sec:data_faq_corpus}
For FAQ retrieval, we annotate each query in the query reformulation test set with a ground-truth FAQ, ensuring full coverage for the test set in our experiments.\footnote{We cannot disclose the performance on the full internal FAQ database.}%

\subsection{Implementation Details for Our Method}\label{subsec:ours}

\noindent \textbf{Intent Classifier}
We fine-tuned RoBERTa large model \citep{liu2019roberta} on the intent classification dataset (cf. \S\ref{sec:data_intent_classifier}). Due to significant class imbalance, we upsampled (with repetition) the question-intent queries until balanced distribution of intents are satisfied (we cannot disclose the exact sampling ratio). Standard cross entropy (CE) loss is adopted. This model is trained for 8 epochs with 2e-6 learning rate and 256 batch size, which is distributed evenly on 8 NVIDIA K80 GPUs. We use Adam as our optimizer and early stopping ($patience=10$) is used to prevent overfitting.\\

\noindent \textbf{Reformulation Model}
We train two reformulation models using BART-base~\cite{DBLP:journals/corr/abs-1910-13461} and T5-base~\cite{2020t5}, respectively. Both models take a user query with question intent as input, and output its reformulated question. The CE loss is adopted to train the models by maximizing the likelihood of generating the human's reformulation. This model is trained for 10 epochs with 1e-5 learning rate and a single Tesla A100 GPU. We use Adam as our optimizer, and batch size of 16. The training is halted using early stopping ($patience=3$). All intent classifier and reformulation models are implemented using HuggingFace.\footnote{\tiny{\url{https://huggingface.co/}}}

\subsection{Implementation Details for FAQ Retrieval}\label{subsec:faq}
We evaluate our approach for FAQ retrieval on the following ranking models of different complexity:
\paragraph{BM25~\cite{robertson2009probabilistic}} We use BM25 as an unsupervised FAQ retrieval model. All the FAQ questions are indexed using Lucene.\footnote{\tiny{\url{https://lucene.apache.org/}}} For a user query, we retrieve top-50 documents based on BM25 scores regardless of the query intent. It is possible that some queries with non-question intent may not return any results.

\paragraph{SentTrans~\cite{reimers-2019-sentence-bert}}
We adopt a sentence encoder model\footnote{\tiny\url{https://www.sbert.net/docs/pretrained-models/nq-v1.html}} that is trained on Google's Natural Question corpus \citep{kwiatkowski2019natural} to compute the similarity between queries reformulations and FAQ questions. For retrieval, we rank FAQ questions based on their cosine similarity against either the query or its rewritten question.

\paragraph{BERT~\cite{devlin-etal-2019-bert}} We fine-tune a pointwise ranker using BERT on our query reformulation dataset to rank a query against all FAQs. For a query, we treat the ground-truth reformulation as the positive sample and randomly sample 100 reformulations of other queries as negative samples. Finally, the FAQ questions are ranked based on the classification score w.r.t the query.
Hinge loss function is applied to train the model for 10 epochs with a batch size of 25.

\paragraph{BERT-Rerank~\cite{dai2019deeper}} Directly ranking all FAQs with BERT is computationally expensive. A more efficient approach is to rerank the top-$k$ results of BM25 using BERT. In our experiments, we test $k$ with 10 and 50.

\subsection{Simulating Intent Classification with Retrieval Baselines}
\label{sec:simulate_intent_classification}
Although the FAQ retrieval approaches discussed in \S\ref{subsec:faq} were not originally designed for intent classification, we include them as baselines to approximate false matches on real traffic. Specifically, we evaluate the probability of FAQ results shown to users who do not have the intent of seeking FAQs (reflected by precision), as well as the likelihood of FAQ results not appearing for users who are specifically searching for FAQs (reflected by recall), in the presence and absence of our intent classifier. For BM25, we consider a query to have question intent if it meets both requirements:

\begin{enumerate}
    \itemsep0em
    \item The number of returned FAQ results is more than a threshold $x$;
    \item BM25 score of the top-1 result is larger than a threshold $y$.
\end{enumerate}

By default, we set $x=1$ and $y=0$, which means at least one FAQ is returned for a question intent query. To obtain optimal BM25 results, we use the validation set for fine-tuning the thresholds ($x=40$, $y=5$). Similarly for SentTrans, we find the optimal cosine similarity threshold ($0.6$) based on the validation set, and classify queries with question intent if an FAQ is retrieved above the threshold.

\subsection{Evaluation Metrics}

To measure the intent classification performance, we report results on precision, recall and F1. 
We report mean reciprocal rank (MRR) and Hit$@1$ to evaluate the FAQ retrieval performance. Hit$@1$ is the most critical metric since only the top-1 retrieved FAQ result will be displayed to users.

\section{Experiments and Results}
\label{sec:rs}

We study the following research questions:
\begin{enumerate}[label={\bfseries RQ\arabic*:},leftmargin=*]
    \itemsep0em
    \item How much does using the question intent classifier as a filtering step benefit the integrated FAQ retrieval pipeline?
    \item How effective is query-to-question reformulation for FAQ retrieval?
    \item How efficient is our intent-aware FAQ retrieval approach when integrated with a product search engine?
\end{enumerate}

\subsection{Question Intent Classification}\label{sec:intent_rs}

The intent classifier determines when to trigger FAQ retrieval and display the FAQ results alongside the product search results. 
Table~\ref{t:intent} shows the evaluation results on intent classification. For reasons of confidentiality, we report results as relative differences with the baseline method (BM25). The details on approximating precision and recall for baselines are summarized in \S \ref{sec:simulate_intent_classification}. Since the evaluation set is highly imbalanced, BM25 has the lowest F1 score with the majority of queries being classified as question intent queries, leading to extremely high recall and low precision. Even with optimal thresholds, BM25 obtains only an F1 score that is 48\% lower than our method. Although SentTrans outperforms BM25, it still falls significantly behind our method. To answer \textbf{RQ1}, intent classifier improves precision by 26\% and recall by 51\% compared to the strongest SentTrans baseline.

\begin{table}[h]
\centering
\resizebox{.95\columnwidth}{!}{
    \begin{tabular}{lrrr}
    \toprule
    \textbf{Methods} & \textbf{Precision} & \textbf{Recall} & \textbf{F1} \\ \midrule
    \textbf{BM25} & 0.00 & \textbf{0.00} & 0.00 \\
    \textbf{BM25 (Optimal)} & +0.41 & -0.60 & +0.36 \\
    \textbf{SentTrans (Optimal)} & +0.54 & -0.54 & +0.46 \\ 
    \textbf{Our Method} & \textbf{+0.80} & -0.03 & \textbf{+0.84} \\ \bottomrule
    \end{tabular}}
    \caption{Results of question intent classification. Scores are relative to BM25.}\label{t:intent}
\end{table}

These results indicate that question intent classifier is required because other baselines cannot effectively distinguish queries with question intent.%

\subsection{FAQ Retrieval}\label{sec:faq}

For each retrieval system being compared, there are three input types: (1) original query, (2) BART reformulated query, and (3) T5 reformulated query. Table~\ref{t:faq} summarizes the FAQ retrieval results. Similar to Section~\ref{sec:intent_rs}, we only report numbers relative to the BM25 baseline where original queries are used (first row in Table~\ref{t:faq}).

\begin{table}[]
\centering
\resizebox{1.0\columnwidth}{!}{
\begin{tabular}{llrr}
\toprule
\textbf{Retrieval Model} & \textbf{Input Query Type} & \textbf{MRR} & \textbf{Hit@1} \\ \midrule
\multirow{3}{*}{BM25} & Original & 0.00 & 0.00 \\
 & Reformulation (BART) & +0.07 & +0.10 \\
 & Reformulation (T5) & +0.11 & +0.17 \\ \midrule
\multirow{3}{*}{SentTrans} & Original & +0.01 & +0.01 \\
 & Reformulation (BART) & +0.08 & +0.12 \\
 & Reformulation (T5) & +0.13 & +0.19 \\ \midrule
\multirow{3}{*}{BERT (pointwise)} & Original & +0.02 & +0.02 \\
 & Reformulation (BART) & +0.04 & +0.06 \\
 & Reformulation (T5) & +0.07 & +0.09 \\ \midrule
\multirow{3}{*}{BERT-Rerank (top-10)} & Original & +0.05 & +0.07 \\
 & Reformulation (BART) & +0.10 & +0.15 \\
 & Reformulation (T5) & \textbf{+0.14} & \textbf{+0.20} \\ \midrule
\multirow{3}{*}{BERT-Rerank (top-50)} & Original & +0.08 & +0.08 \\
 & Reformulation (BART) & +0.10 & +0.13 \\
 & Reformulation (T5) & \textbf{+0.14} & +0.19 \\ \bottomrule
\end{tabular}}
\caption{Results of FAQ retrieval by different retrieval models and query types. Scores are relative to BM25 using the original queries.}\label{t:faq}
\end{table}

First, we observe that using the reformulated query improves the performance of all FAQ retrieval systems in terms of MRR and Hit$@1$, which answers \textbf{RQ2}. T5 query reformulations consistently show better results than BART query reformulations. %
Even for the most competitive BERT-Rerank baselines,  T5 can further improve Hit$@1$ by more than 10\%.
The results indicate that our reformulation method has the advantage of improving the precision at top ranks. 
Although our reformulation models are not designed to generate exact questions from FAQs,\footnote{This is a challenging annotation task because workers must match queries against our entire FAQ corpus.} they generate questions that are sufficiently similar to the FAQs in our corpus. As a result, they significantly improve the retrieval performance.

Second, Table~\ref{t:faq} shows that reformulations allow BM25 to achieve comparable results with strong BERT-Rerank baselines. Notably, BM25 with T5's reformulated queries outperforms two BERT-Rerank baselines using the original queries. Yet, when using T5's reformulations, BERT-Rerank (top-10) achieves the best MRR and Hit$@1$ among all methods, outperforming the original query by 13\% in Hit$@1$, and 9\% in MRR.
Considering the model's complexity, the combination of BM25 and the reformulation method is a promising FAQ solution in industry settings.

\subsection{Efficiency Comparison}
\label{sec:efficiency}

An important consideration in industry settings is the efficiency of the proposed solution. We assess the impact of our proposed solution in terms of computational cost. As BERT Rerank (top-10) with T5 reformulation yields the best results, we compare its inference time with and without an intent classifier. We measure the speed on a single \emph{p3.8xlarge} instance,\footnote{\tiny{\url{https://aws.amazon.com/ec2/instance-types/}}} with a batch size of 16.

Table~\ref{t:time} shows the inference time\footnote{The raw inference time cannot be disclosed.} normalized by the baseline that does not use an intent classifier. For \textbf{RQ3}, our proposed intent-aware FAQ retrieval system that reranks top-10 results using BERT and T5 reformulations can save on average 95\% of inference time, which is a significant benefit for real-world applications. We also observed that despite adding an extra layer of inference from reformulation model, the increase in latency is negligible since majority of queries are already filtered out and only a small fraction of queries are reformulated.

\begin{table}[h]
\centering
\resizebox{1.0\columnwidth}{!}{
\begin{tabular}{llr}
\hline
\textbf{Pipeline} & \textbf{Query} & \textbf{Inference Time} \\ \hline
BERT-Rerank (top 10) & Original & 1.0 \\ \hline
\multirow{3}{*}{Ours} & Original & 0.0404 ($\downarrow$95.96\%) \\
 & Reformulation (BART) & 0.0446 ($\downarrow$95.54\%) \\
 & Reformulation (T5) & 0.0461 ($\downarrow$95.39\%) \\ \hline
\end{tabular}}
\caption{Inference speed comparison on full query traffic (including question intent and non-question intent queries) with and without intent classifier. The inference time is normalized based on the time taken by the method that does not utilize an intent classifier.} \label{t:time}
\end{table}

\section{Online Deployment}

Our intent-aware FAQ retrieval solution is deployed and integrated into the product search interface of a leading global e-commerce website.
The live version uses a much larger FAQ corpus than the ones used here.
When FAQ results are displayed, we collect optional explicit user feedback on whether the answer is helpful or not, as demonstrated in Figure~\ref{fig:faq_example}. 
Over one month of traffic was collected from the US marketplace, showing that 71\% of the rendered FAQ results received explicit positive customer feedback. The feedback results were aggregated at the query level, allowing each query to receive multiple positive feedback responses. %
These findings demonstrate that our practical solution is not only effective in offline evaluation, but also helpful to real users.

%% file: conclusion.tex
\section{Conclusion}

We proposed to integrate FAQ retrieval with product search to address challenges in aggregated search from an e-commerce perspective. Our approach first classifies queries with question intent, which then reformulates into natural language questions that are used to retrieve FAQs. Offline experimental results show that on the best-performed FAQ retrieval system (\ie BERT-Rerank (top 10)), the proposed intent classifier saves a substantial amount of inference costs (95\%) and also improved retrieval performance through query reformulation by 13\% on Hit$@1$. These improvements are also reflected in online evaluation: over one month of user feedback demonstrated that about 71\% of the rendered FAQ results were considered to be helpful. Overall, the findings in this work suggest promising directions for e-commerce platforms to support FAQ retrieval at scale without disrupting customer's shopping experience.

%% file: limit.tex
\section*{Limitations and Future Work}

One limitation of our approach is that we do not display or rank multiple reformulations. It is possible that a query can be reformulated into multiple possible questions. For example, the query \example{apple tv bluetooth} can be reformulated into \example{How do I connect a Bluetooth device to my Apple TV} or \example{Does apple TV support Bluetooth}. In our future work, we aim to explore the integration of multiple reformulations into the FAQ retrieval process to further enhance the overall user experience. Another limitation is that we do not train an end-to-end FAQ retrieval model. In the future, we plan to train the FAQ retrieval model using the reformulations so that the original query can directly be used.